\documentclass[12pt]{article}
\usepackage{spconf,amsmath,epsfig}
\usepackage{tikz}
\usepackage{multirow}
\usepackage{makecell}
\usepackage{pgfplots}
\pgfplotsset{compat=1.17}
\usetikzlibrary{shapes.geometric, arrows}
\usepackage{tcolorbox}
\tcbuselibrary{breakable}

\usepackage[a4paper,margin=2cm]{geometry}

\tikzstyle{process} = [rectangle, rounded corners, minimum width=3cm, minimum height=1cm, text centered, draw=black, fill=blue!20]
\tikzstyle{arrow} = [thick,->,>=stealth]
\usepackage{lipsum}
\usepackage{booktabs}
\usepackage{subcaption}
\usepackage{rotating}
\usepackage{tcolorbox}
\usepackage{enumitem}
\usepackage[colorinlistoftodos,prependcaption,textsize=normalsize]{todonotes}
\usepackage{float}
\usepackage[commandnameprefix=always]{changes}
\tcbuselibrary{listingsutf8}
\providecommand{\doi}[1]{doi: {\footnotesize \href{http://dx.doi.org/#1}{\path{#1}}}}

\usepackage[pdftex=true,breaklinks=true,hidelinks=true,colorlinks=true,citecolor=blue]{hyperref}

\usepackage[square,numbers]{natbib}

\title{Beyond detection: cooperative multi-agent reasoning for rapid onboard EO crisis response}
\name{
\begin{tabular}{c}
Alejandro D. Mousist, Pedro Delgado de Robles Martín \\
Raquel Lladró Climent, Julian Cobos Aparicio
\end{tabular}
}

\address{Thales Alenia Space España\\
C/ Einstein 7 (PTM)\\
28760 Tres Cantos, Madrid, Spain\\
alejandro.mousist@thalesaleniaspace.com\\
pedro.delgado@external.thalesaleniaspace.com\\
raquel.lladrocliment@thalesaleniaspace.com\\
julian.cobos@thalesaleniaspace.com}

\begin{document}
\raggedbottom
%
\maketitle
%
\begin{abstract}
Rapid identification of hazardous events is essential for next-generation Earth Observation (EO) missions supporting disaster response. However, current monitoring pipelines remain largely ground-centric, introducing latency due to downlink limitations, multi-source data fusion constraints, and the computational cost of exhaustive scene analysis.

This work proposes a hierarchical multi-agent architecture for onboard EO processing under strict resource and bandwidth constraints. The system enables the exploitation of complementary multimodal observations by coordinating specialized AI agents within an event-driven decision pipeline. AI agents can be deployed across multiple nodes in a distributed setting, such as satellite platforms. An Early Warning agent generates fast hypotheses from onboard observations and selectively activates domain-specific analysis agents, while a Decision agent consolidates the evidence to issue a final alert. The architecture combines vision–language models, traditional remote sensing analysis tools, and role-specialized agents to enable structured reasoning over multimodal observations while minimizing unnecessary computation.

A proof-of-concept implementation was executed on the engineering model of an edge-computing platform currently deployed in orbit, using representative satellite data. Experiments on wildfire and flood monitoring scenarios show that the proposed routing-based pipeline significantly reduces computational overhead while maintaining coherent decision outputs, demonstrating the feasibility of distributed agent-based reasoning for future autonomous EO constellations.
\end{abstract}

%

\section{INTRODUCTION}
\label{sec:intro}
Despite the increasing number of EO satellites, improved revisit times, and the growing availability of heterogeneous sensors, the operational pipeline for hazard monitoring remains largely ground-centric and bandwidth-limited. As a result, many observations are either downlinked with significant delay or never exploited for time-critical decision making.

Timely and reliable detection of hazardous events is therefore a key requirement for next-generation Earth Observation (EO) missions supporting emergency response and risk mitigation, particularly in the case of floods and wildfires. Despite increasing revisit capabilities, operational latency remains dominated by three main bottlenecks: limited downlink capacity, the computational cost of analyzing acquired scenes, and the latency introduced by human-in-the-loop acquisition workflows. These limitations have motivated a shift toward onboard processing, enabling earlier alert generation directly in space \cite{eo-alert}.

A key challenge is how to reliably detect and prioritize hazardous events directly onboard using heterogeneous sensing modalities. Multimodal observations provide complementary information, but each modality presents specific limitations. Optical imagery offers rich spatial and spectral detail but is affected by clouds and illumination. Thermal data highlights temperature anomalies, while radar (e.g., SAR) operates independently of daylight and weather conditions.

However, performing full multimodal processing on every observation can be infeasible onboard due to power and compute constraints. Furthermore, both integrating multiple sensors on a single satellite and coordinating observations across satellites introduce significant system-level complexity.

Several approaches have attempted to mitigate these limitations through onboard processing or distributed observation strategies. Early-warning pipelines based on lightweight algorithms can reduce downlink requirements but typically rely on single-modality analysis and fixed decision rules. More recent agentic AI systems enable flexible geospatial reasoning using large language models and domain-specific tools, yet they are generally designed for cloud environments rather than operating under realistic onboard resource constraints.

To address these limitations, this work proposes a hierarchical distributed architecture in which nodes assume differentiated functional roles within an event-driven processing pipeline. Incoming observations are first screened through lightweight hypothesis generation mechanisms, triggering deeper and more specialized analyses only when necessary. Analytical outputs from heterogeneous tools are subsequently reconciled and fused to issue a final alert. This event-driven design reduces unnecessary computation, enables horizontal scalability, and enhances robustness through multimodal diversity.

Crucially, this differentiation is not merely structural but cognitive: some agents generate rapid, structured hypotheses using Vision-Language Models (VLMs), while others leverage Large Language Models (LLMs) to consolidate the results of more exhaustive machine-learning and traditional remote-sensing analyses, as well as to perform decision-level evidence fusion. Extending previous work on language-model–driven autonomous reasoning for spaceborne systems, this study advances the paradigm toward a distributed, role-specialized multi-agent swarm.

To assess the feasibility of this approach without requiring full swarm-level deployment, we validate the proposed decision framework through a controlled multi-agent demonstrator executed on the engineering model of an edge-computing platform currently deployed in orbit. The experimental setup emulates distributed satellite agents while preserving the hierarchical and event-driven structure of the envisioned swarm-based system. Operating under hardware conditions representative of real onboard constraints, the demonstrator enables reproducible evaluation of routing efficiency and decision-level evidence fusion in a realistic spaceborne computing environment.
\medskip

The main contributions of this work are:
\begin{enumerate}
    \item A hierarchical, event-driven distributed system architecture for rapid onboard multimodal hazard detection, exploiting distributed multimodal observations.
    \item The use of language-based models within a multi-agent Earth Observation pipeline to support hypothesis generation, analytical consolidation, and final decision making.
    \item The validation of the proposed decision framework through a controlled and representative multi-agent implementation executed on the engineering model of an edge-computing platform currently deployed in orbit.
\end{enumerate}

\section{Related Work}
\label{sec:results}
Recent technology demonstrations have explored distributed autonomy in satellite swarms. NASA’s Starling mission validated key capabilities for cooperative spacecraft clusters, including inter-satellite networking, onboard decision-making, and autonomous maneuver planning within a swarm of CubeSats \cite{starling}. These experiments illustrate the feasibility of coordinating multiple spacecraft as distributed systems capable of executing autonomous operations in orbit.

The rapid growth of EO satellite constellations has further reinforced this trend, transforming modern missions into inherently distributed systems involving satellites, ground stations, operators, and users. Coordinating these assets involves complex planning and resource allocation problems under dynamic conditions. As noted by Picard et al. \cite{multiagent}, such challenges naturally align with the multi-agent paradigm, where multiple autonomous entities cooperate and adapt to operate the constellation efficiently.

Federated EO systems extend this concept by coordinating multiple satellite missions in order to distribute acquisition tasks across heterogeneous constellations and increase the temporal and modal coverage of a target area \cite{federation}. Such approaches enable the combination of complementary sensing modalities from different missions, potentially providing richer multimodal observations of a region of interest. However, these systems primarily focus on observation planning and data sharing across missions, and typically assume that the target area is known in advance. The integration and interpretation of the resulting observations are still largely performed on the ground.

Current approaches in next-generation EO multi-sensor constellations, such as the Atlantic Constellation \cite{atlantic_constellation} and IRIDE \cite{iride}, typically address these challenges by simplifying onboard processing and delegating multimodal data fusion to the ground segment. In the first case, onboard algorithms often rely on a single sensor modality and simple rule-based processing, reducing computational load but limiting robustness under adverse conditions such as cloud cover, smoke, or complex land cover. In the second case, multimodal analysis is postponed until all data are received on the ground, introducing additional delays and limiting the ability to dynamically prioritize observations.

In parallel, recent work has explored the use of agentic artificial intelligence to enable flexible and tool-augmented reasoning in remote sensing workflows. 
Recent studies have also investigated the use of LLM-based agents for autonomous spacecraft operations, including tasks such as system monitoring and thermal control \cite{astrea}.
These results suggest that language-model–driven reasoning can be applied to onboard decision-making problems.
Agentic AI has been proposed as a paradigm shift from static perception pipelines toward autonomous, tool-augmented, and goal-directed decision-making in remote sensing \cite{agentic_ai_survey}. These systems combine large language models with domain-specific tools to enable interactive reasoning and flexible geospatial analysis.

In Earth-Agent \cite{earth-agent}, the authors emphasize the necessity of equipping LLM-based agents with a structured ecosystem of domain-specific tools to enable scientifically grounded geospatial analysis beyond RGB perception, extending toward spectral and product-level reasoning. Similarly, RS-Agent \cite{rs-agent} integrates large language models with remote sensing analysis tools to support interactive EO reasoning tasks driven by user queries. Systems such as NAIAD \cite{naiad} explore tool-augmented agents for environmental monitoring workflows combining satellite observations with additional data sources.

Despite these advances, most existing agentic EO systems are designed for cloud-based environments and interactive analytical workflows executed on the ground. As a result, they typically assume abundant computational resources and do not explicitly address the constraints of spaceborne platforms.

Bringing agentic reasoning capabilities onboard therefore requires system-level architectures that combine distributed sensing, hierarchical processing, and event-driven decision pipelines capable of operating within strict onboard resource constraints. This gap motivates the distributed multi-agent architecture proposed in this work, which integrates early hypothesis generation, specialized multimodal analysis, and decision-level evidence fusion directly within the observation pipeline.

\section{Proposed System Architecture}
\label{sec:general_design}

To address these limitations, we propose a distributed multi-agent architecture in which AI agents operate directly onboard to enable hierarchical and event-driven processing of multimodal
EO observations.

\subsection{Architectural Rationale}
Ideally, each scene would be observed using multiple complementary sensors, combining optical, thermal, and radar information to improve reliability and reduce ambiguity.

In practice, performing a full multimodal analysis of every observed scene is computationally prohibitive under onboard resource constraints, particularly in time-critical scenarios. Additionally, integrating heterogeneous instruments such as SAR, hyperspectral, and thermal infrared sensors on a single small satellite introduces significant platform-level constraints. These sensors have different power, thermal, structural, and pointing requirements, which complicate system design and often force performance compromises. Co-locating them also increases subsystem coupling, meaning that a failure in one component may degrade overall mission performance.

\subsection{Hierarchical Distributed Design}

To address the limitations described before, this study proposes a hierarchical distributed architecture. Distributing sensors across multiple interconnected satellites in a distributed architecture allows each platform to be optimized for a specific sensing modality, preserving design flexibility and reducing resource coupling.

At the same time, processing is organized in layers: early-stage nodes perform lightweight screening to identify candidate events, while deeper, computationally intensive analysis is only triggered when necessary. This layered approach enables early rejection of non-threatening scenes, conserving onboard resources for cases that require exhaustive assessment.

\begin{figure}[ht]
  \centering
  \includegraphics[width=\columnwidth]{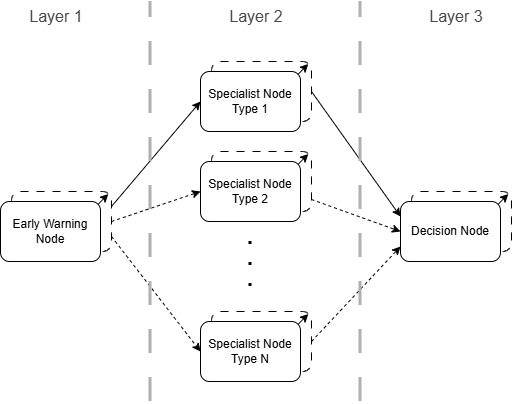}
  \caption{Event-driven hierarchical architecture activated upon candidate event detection.}
  \label{fig:general_design}
\end{figure}

Fig \ref{fig:general_design} illustrates the event-driven data flow triggered when Early Warning nodes detect a candidate event. 

\subsection{Observational Diversity and Role-Based Redundancy}

An additional advantage of this distributed approach is observational heterogeneity. Different satellites may observe the same region from distinct viewing geometries, or at slightly different times depending on orbital configuration. As a result, conclusions are derived from diverse perspectives rather than a single acquisition context. Detection decisions therefore rely on consistency across independent observations, increasing robustness and reducing the likelihood of false alarms.

Each satellite in the swarm-based system is assigned a well-defined sensing capability and corresponding operational role within the fleet.
Functional specialization has also been explored in recent multi-agent geospatial systems, where orchestration is separated from domain-specific analysis and tasks are delegated to expert agents equipped with dedicated toolsets. Such architectures have been shown to scale more effectively as the number of remote-sensing workflows and analytical tools increases \cite{geollm_squad}. This functional specialization enables role-based redundancy: if a given node fails, it can be replaced or reinforced by another satellite carrying the same sensing modality and responsibilities. While the replacement node may not occupy the exact same orbital position at a given instant, it preserves the functional role within the cluster.

Rather than representing a limitation, this orbital non-coincidence can provide additional observational diversity, introducing alternative viewing geometries or acquisition times over the same event. Such diversity further strengthens detection reliability by incorporating complementary perspectives into the decision process.

This decoupling between physical position and functional responsibility enhances system resilience and scalability. The swarm-based system can be incrementally expanded by adding additional specialist nodes, and localized failures do not compromise the overall architecture, but only temporarily affect coverage or revisit performance for that specific modality.

\begin{figure}[ht]
  \centering
  \includegraphics[width=\columnwidth]{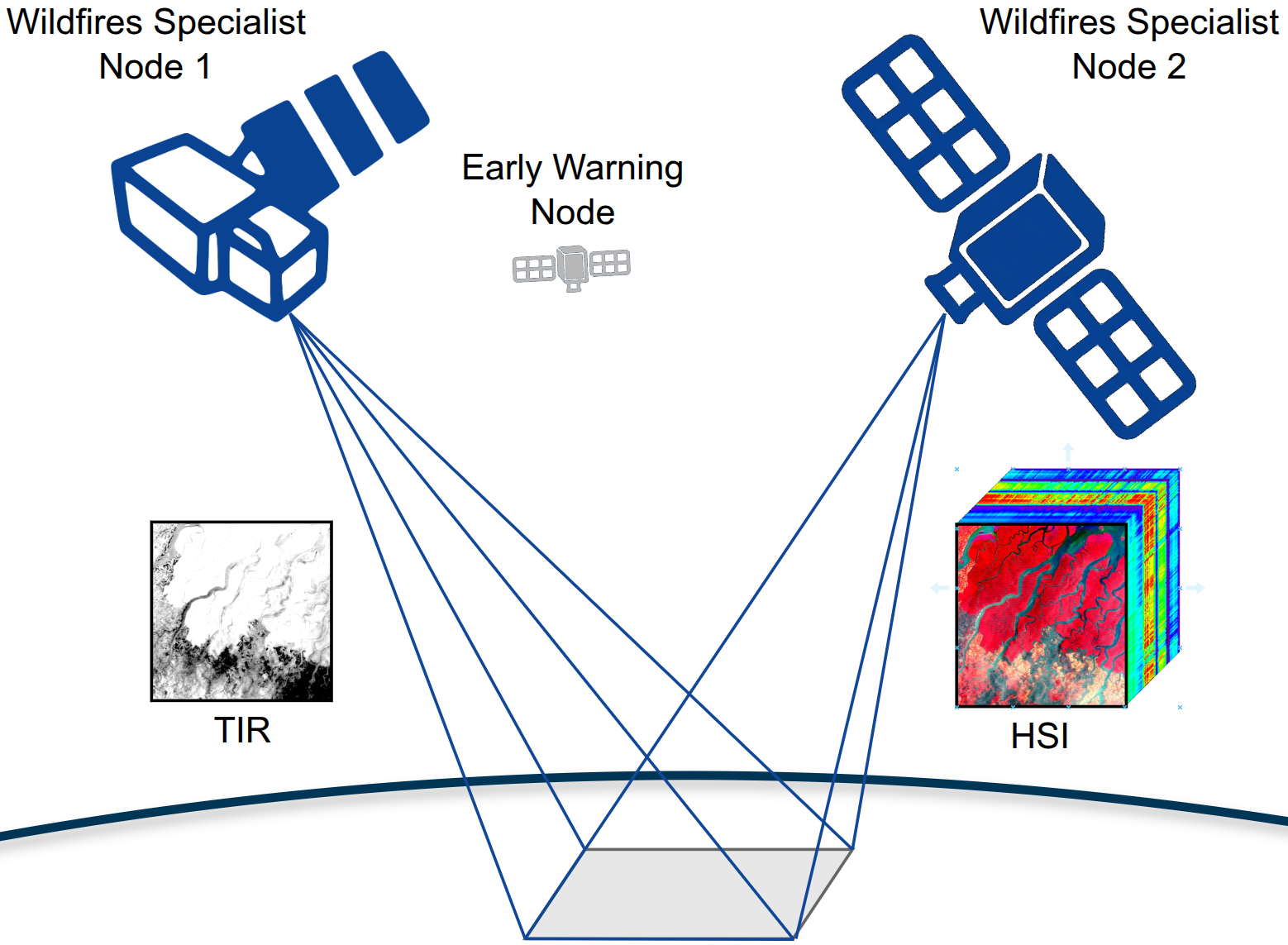}
  \caption{Two wildfire specialist nodes with different sensing modalities (TIR and HSI) observing the same area of interest after candidate wildfire detection by the Early Warning node.}
  \label{fig:multisat}
\end{figure}

Fig. \ref{fig:multisat} illustrates two specialist satellites fulfilling the same functional role, wildfire detection, using different sensing modalities. While each platform remains optimized for a specific instrument, functional equivalence is defined at the analytical level rather than at the sensor level.

\subsection{Event-Driven Processing and Scalability}

The proposed architecture follows a unidirectional, event-driven flow. Processing is initiated by the Early Warning layer, which continuously monitors observed scenes using lightweight screening mechanisms. When a potential threat is identified, the corresponding specialist nodes are selectively activated, and their structured outputs are subsequently forwarded to the Decision layer for final assessment and alert generation.

This directional pipeline enables horizontal scalability at each layer. Increasing the number of Early Warning nodes enhances area coverage and reduces detection latency. Expanding the pool of specialist nodes allows parallel confirmation of multiple candidate events. Similarly, multiple Decision nodes can operate concurrently to handle incoming evidence streams and issue timely alarms.

\section{Reference Implementation}
\label{sec:implementation}
To validate the architecture proposed in Chapter~\ref{sec:general_design}, a scenario involving a four-satellite fleet arranged in a diamond configuration was implemented. This demonstrator constitutes a concrete instantiation of the distributed multi-agent architecture described previously, preserving its hierarchical decision structure while constraining execution to a controlled experimental setup.

For experimental simplicity, the demonstrator employs an external workflow orchestrator to coordinate agent execution. In an operational deployment, the processing flow would be event-driven and triggered directly by Early Warning nodes upon detection of candidate threats, thereby preserving the distributed nature of the architecture.

\begin{figure}[ht]
  \centering
  \includegraphics[width=\columnwidth]{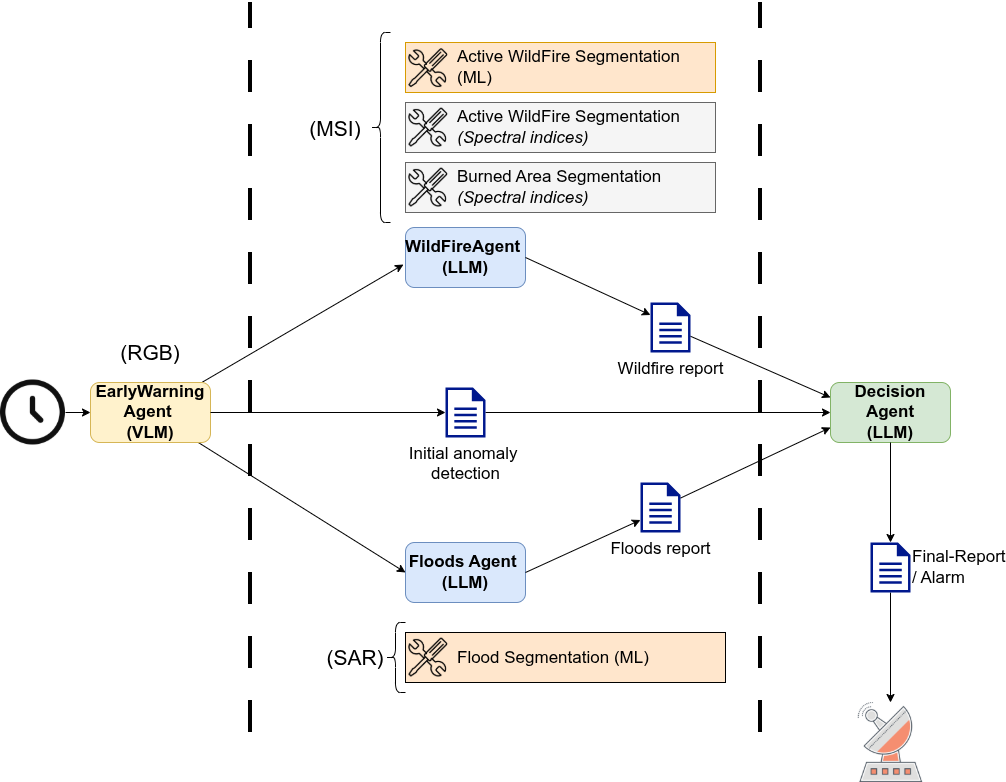}
  \caption{Diamond-Topology Multi-Agent Demonstrator and information flow between agents.}
  \label{fig:reference_impl}
\end{figure}

In this configuration, an initial node, referred to as the \emph{Early Warning Agent}, performs a preliminary analysis of an RGB image of a scene to detect potential floods or active wildfires. This stage provides a coarse perception layer that generates an initial hypothesis regarding the presence of a hazardous event. When an event is suspected, the corresponding specialist nodes (\emph{Wildfires Specialist Agent} and/or \emph{Flood Specialist Agent}) are activated to conduct a more detailed domain-specific analysis using advanced tools focused on the area of interest.

Subsequently, a fan-in stage is performed at a final node, the \emph{Decision Agent}, which receives both the initial hypothesis and the structured reports produced by the specialist agents. Acting as a fusion module, the \emph{Decision Agent} integrates these sources of evidence, resolves potential inconsistencies between them, and determines whether a threat is present. It then issues a final report explicitly stating its decision and the rationale derived from the preceding analyses.

The objective of this implementation is not to validate orbital coordination or realistic inter-satellite communication protocols, but rather to assess the functional behavior of the multi-agent decision pipeline. The primary focus is placed on the agents’ ability to interpret intermediate results, their response to inconsistencies, and their performance on representative hardware.

To limit the scope, the processing flow is statically defined within a workflow externally orchestrated by a fifth node using Prefect\cite{prefect_open_source} as the workflow manager. Additionally, inter-node communication is simplified to synchronous connections implemented through REST endpoints. Each agent operates within an independent container that simulates a physical node. The implementation abstracts the communication layer to focus on decision logic, while preserving the architectural constraints and interaction patterns expected in a distributed deployment.

\subsection{Early Warning Agent}

The \emph{Early Warning Agent} provides an initial assessment by analyzing incoming RGB scenes derived from Sentinel-2 (S2) MSI data in the context of the proof-of-concept, producing a rapid hypothesis that guides the subsequent invocation of specialist agents. It relies on an instruction-tuned Qwen2-VL \cite{qwen2vl} vision-language model (VLM) with 2 billion parameters, quantized to 4-bit precision for efficient on-board inference.

When tasked with assessing potential hazards in a scene, the VLM is constrained to produce a short, structured response indicating the suspected disaster type along with a concise semantic explanation, encoded in JSON format. This structured output enables downstream agent coordination, traceability, and explainable decision fusion.

\subsection{Wildfires Specialist Agent}
The \emph{Wildfires Specialist Agent} performs a detailed analysis of S2 MSI scenes to detect fire events, primarily exploiting information from the SWIR1 (B11), SWIR2 (B12), and NIR (B8) spectral bands. The outputs produced by the specialized tools are then interpreted by a Qwen-2.5\cite{qwen2.5} instruction-tuned large language model with 3 billion parameters, quantized to 8-bit precision, which generates a structured JSON output summarizing the detection results and providing a semantic explanation of the identified wildfire patterns.
The model sizes were empirically chosen to balance computational efficiency and reasoning performance across the pipeline.

\subsubsection{Tools}
\begin{figure}[ht]
  \centering
  \includegraphics[width=\columnwidth]{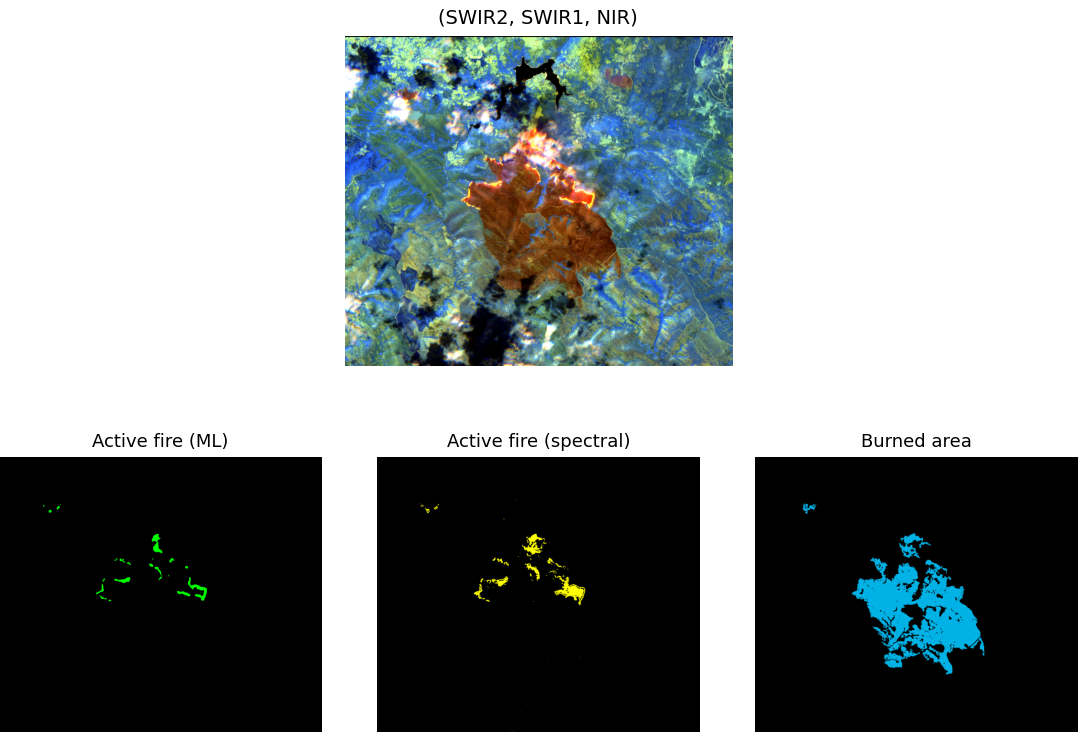}
  \caption{(B12, B11, B8) composite from a Sentinel-2 scene used to enhance the visualization of active wildfire signatures. The segmentations produced by the three proposed methods are shown for the same scene.}
  \label{fig:wildfire_tools}
\end{figure}

The agent employs three tools for wildfire detection using S2 MSI imagery. Two of these tools focus on active wildfires, combining machine learning and spectral indexes; depending on the fire stage, they may detect either the entire wildfire during early phases or only the fire perimeter at later stages. The third tool targets burned areas and relies exclusively on spectral indexes. Fig. \ref{fig:wildfire_tools} shows a Sentinel-2 scene where previously burned areas are located in the central region, while active fire fronts are visible along the perimeter of the burned scar.

\begin{enumerate}
  \item \textbf{Active wildfire segmentation using machine learning (ML).}
  The tool identifies active fires by using a ML model built on a DeepLabV3+ \cite{deeplabv3plus} architecture with a ResNet-50 \cite{resnet50} backbone that performs semantic segmentation in the image. 
  
  The model was trained to detect active wildfires from S2 MSI images using the SWIR and NIR spectral response. It takes as input a three band 256 × 256 pixel image formed by S2 bands B8 (NIR), B11 (SWIR1), and B12 (SWIR2). This combination is known to capture the characteristic spectral behavior of active fires (high response in SWIR) and the surrounding land cover (better represented in NIR), which helps the model to discriminate active hotspots from background surfaces. 
  
  The training set is composed of two parts: (i) a subset of image patches extracted from eleven S2 scenes containing wildfire events, and (ii) patches from the Sen2Fire dataset \cite{sen2fire}. All patch labels were generated using a U-Net-based model described in \cite{transfer_learning_tf}. The tool outputs a fire/no-fire decision along with an estimate of the active fire area.

  \item \textbf{Active wildfire segmentation using indexes.}
  The tool is able to segment active fires in S2 by calculating the Normalized Hotspot Indexes (NHI) \cite{nhi} which consists of two indexes that complement each other: NHISWIR (See eq. \ref{eq:nhi_swir}) and NHISWNIR (See eq. \ref{eq:nhi_swnir}).
  NHI indexes are defined as:
  \begin{equation}
      NHI_{SWIR} = \frac{\rho_{SWIR2} - \rho_{SWIR1}}{\rho_{SWIR2} + \rho_{SWIR1}}
      \label{eq:nhi_swir}
  \end{equation}
  \begin{equation}
      NHI_{SWNIR} = \frac{\rho_{SWIR1} - \rho_{NIR}}{\rho_{SWIR1} + \rho_{NIR}}
      \label{eq:nhi_swnir}
  \end{equation}

  These NHI indexes exploit spectral contrast in the SWIR and NIR regions and are computed using the corresponding S2 MSI bands (B12, B11, and B8). Although originally formulated using top-of-atmosphere radiance, surface reflectance (bottom-of-atmosphere) from S2 L2A products is used in this study. This choice helps reduce the domain gap with respect to the vision–language model, which was trained on general-purpose RGB imagery, and ensures consistency with the wildfire model training, which relies on bottom-of-atmosphere reflectance. The proposed architecture, however, remains applicable to TOA data in operational scenarios.
  The NHI indexes were observed to generate false positives over water bodies; therefore, water areas were excluded by applying a mask derived from the Modified Normalized Difference Water Index (MNDWI)\cite{mndwi} (See eq. \ref{eq:mndwi}).
  \begin{equation}
    MNDWI = \frac{\rho_{Green} - \rho_{SWIR1}} {\rho_{Green} + \rho_{SWIR1}}
    \label{eq:mndwi}
  \end{equation}
  In eq. \ref{eq:mndwi}, $\rho_{Green}$ and $\rho_{SWIR1}$ correspond to S2 MSI bands B3 and B11, respectively. 
  
  \item \textbf{Burned areas segmentation using indexes.}
  First, the tool derives a permissive candidate mask following the same formulation as the active-fire index tool described in the previous step, but uses more relaxed thresholds to maximize sensitivity and retain potential fire-related candidates. The candidate mask is then refined using the Burned Area Index (BAI) (\ref{eq:bai}), keeping only pixels supported by both cues.
  \begin{equation}
    BAI = \frac{1}{\left(0.1 - \rho_{Red}\right)^2 + \left(0.06 - \rho_{NIR}\right)^2}
    \label{eq:bai}
\end{equation}

 Since each cue alone can be affected by confounding surfaces and noise, enforcing their intersection provides a simple double-evidence filter that reduces false positives. The tool outputs an estimate of the burned area and the number of hotspots.
\end{enumerate}

\subsection{Flood Specialist Agent}

The \emph{Flood Specialist Agent} follows a processing pipeline similar to that of the wildfire-detection agent but relies on SAR imagery from Sentinel-1 (S1) rather than optical data. Due to the ability of SAR sensors to operate independently of cloud cover and illumination conditions, the analysis relies on the two available polarizations (VV and VH) and on their ratio to enhance sensitivity to water-covered surfaces. These backscatter-based features are particularly effective for detecting inundated areas due to the characteristic low radar return of calm water surfaces. As with the wildfire agent, the outputs of the specialized tools are subsequently processed by the same instruction-tuned large language model to produce a structured JSON report containing the flood/no-flood classification and a corresponding semantic explanation. For simplicity, a shared language model is used across specialist agents in this proof-of-concept, although the architecture supports deploying task-specific models at each node in operational scenarios.

\subsubsection{Tools}
The \emph{Flood Specialist Agent} relies on a single machine learning–based tool for semantic segmentation of flooded areas. This model follows the same semantic-segmentation architecture as the previous Wildfire detection tool (DeepLabV3+ architecture with a ResNet-50 backbone) and is specifically designed for flood mapping using S1 SAR data. 

The model was trained using flood-event samples from the SenForFloods\cite{senforflood} dataset, leveraging S1 imagery acquired during flooding scenarios. The input features consist of the VV and VH polarizations and their ratio (VV/VH), while the reference labels include three classes: flooded areas, permanent water bodies, and dry land.

For both training and inference, S1 scenes are subdivided into overlapping 256 × 256 pixel tiles using a sliding-window approach with a fixed stride. During training, the three-class segmentation scheme was retained to explicitly force the model to discriminate between permanent water surfaces and temporally inundated areas. In the post-processing stage, the segmentation output is subsequently merged into a binary classification distinguishing flooded and non-flooded regions, which is used by the flood-detection agent for decision making using a fixed threshold, where permanent water bodies are included in the non-flooded class.

\begin{figure}[ht]
  \centering
  \includegraphics[width=\columnwidth]{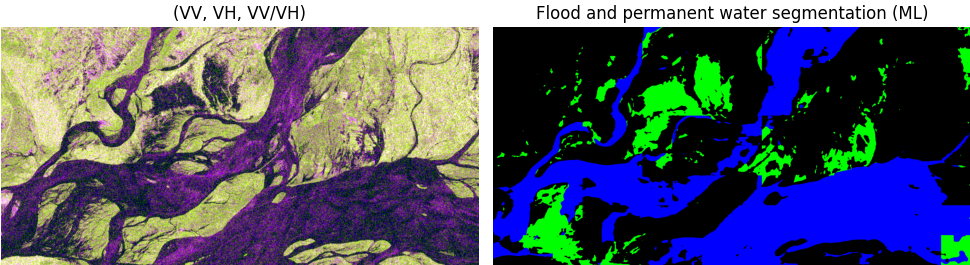}
  \caption{Segmentation of floods and permanent water using the ML model, achieving an IoU$_{0.50}^{flood}$ of 0.554 on the CEMS dataset from SenForFlood. In the segmentation mask, permanent water and flood water are shown in blue and green, respectively.}
  \label{fig:schema}
\end{figure}

\subsection{Decision Agent}
The \emph{Decision Agent} is responsible for producing the system’s final verdict based on the evidence provided by the other agents. To this end, it integrates two main sources of information: (i) the initial hypothesis generated by the Early Warning Agent and (ii) the structured reports produced by the specialist agents.

Although it relies on the same language model as the specialists, its function differs substantially. Rather than interpreting the output of a single tool, it operates as a decision-level fusion module. In particular, it reconciles potential inconsistencies between the initial assessment and the specialist evidence, which may confirm, refine, or contradict the early detection, and generates a final output that is both clear and concise. This output includes the final event type, a confidence score reflecting the level of agreement across sources, and a semantic explanation summarizing the key determining factors.

\section{Experimental Results}
\label{sec:results}
This section evaluates the proposed multi-stage architecture in terms of both computational efficiency and qualitative system behavior. Rather than benchmarking the performance of the underlying disaster-detection models, the experiments focus on assessing the behavior of the hierarchical decision pipeline, particularly the impact of early-stage hypothesis generation on routing efficiency and decision consistency.

To this end, we compare a routing-based execution strategy against a static configuration that always executes all available specialist modules.

Two aspects are analyzed: (i) end-to-end execution time, and (ii) the structure and coherence of the semantic explanations generated from the available evidence. In the baseline configuration, both specialists are executed for every sample and their outputs are consolidated into a single explanation by the onboard language model. In contrast, the proposed pipeline introduces an Early Warning stage that selectively activates only the relevant specialist before issuing the final report.

\subsection{Experimental setup}

The system was evaluated on a curated dataset of 27 samples. Each sample consists of paired Sentinel-2 MSI optical imagery and Sentinel-1 SAR GRD data covering the same scene, labeled as \textit{wildfire}, \textit{flood}, or \textit{no-disaster}.

All experiments were executed on the engineering model of the IMAGIN-e edge computing platform \cite{IMAGINe}, which is currently deployed aboard the International Space Station (ISS). The platform is based on an ARM architecture with 16 CPU cores and 32\,GB of RAM. This hardware configuration provides a representative onboard computing environment for in-orbit inference and autonomous decision-making.
\medskip

Two execution configurations were evaluated:
\begin{itemize}
    \item \textbf{Baseline:} For each sample, both the wildfire and flood specialists are executed with their corresponding tools. The resulting outputs are then consolidated into a final explanation generated by the onboard language model.
    \item \textbf{Proposed pipeline:} The system first executes the Early Warning stage to predict the event class. Based on this result, only the relevant specialists are activated. Subsequently, a final explanation is generated from the Early Warning hypothesis and the structured report/s of the selected specialist/s.
\end{itemize}
\medskip

To isolate its contribution, the Early Warning agent is also independently analyzed to quantify its impact on routing efficiency within the proposed configuration.

\subsection{Performance evaluation}
\label{sec:performace}
We first compare the end-to-end execution time of the baseline and the proposed layered processing pipeline. The baseline always executes both specialists for every sample, whereas the proposed approach includes an early warning stage to selectively activate only the relevant specialist.

As shown in Table \ref{tab:runtime}, the proposed approach yields substantial speed-up in non-disaster scenarios, where unnecessary specialist execution is avoided. In these cases, the computational savings are significant. For event samples (wildfire or flood), where at least one specialist must be executed in both configurations, the improvement is more moderate. In small scenes, the overhead introduced by the early warning stage may partially offset the routing benefit, leading to reduced gains and, in some cases, slightly lower performance compared to the baseline. 

Nevertheless, in real-world scenarios, hazardous events represent a small fraction of observed scenes, while the vast majority correspond to normal (non-disaster) conditions. Consequently, this suggests that overall system-level efficiency gains could be higher in realistic scenarios, where most observations do not require specialist execution.

\begin{table}[H]
\fontsize{10}{12}\selectfont
\centering
\caption{Average speed-up (mean $\pm$ standard deviation) measured on the testbench for 27 samples, grouped by event presence}
\label{tab:runtime}
\begin{tabular}{lcc}
\toprule
\textbf{Event} & \textbf{Speed-up} & \textbf{Reduction (\%)} \\
\midrule
No event & $4.78 \pm 2.54$ & $73.2 \pm 14.1$ \\
\makecell[l]{Event (wildfire \\or flood)} & $1.3 \pm 0.45$ & $13.5 \pm 30.8$ \\
\bottomrule
\end{tabular}
\end{table}

Although the global correlation between scene area and speed-up is negligible ($\rho = 0.08$), a stratified analysis reveals a strong positive correlation in both regimes ($\rho =0.99$ for non-disaster samples and $\rho = 0.92$ for event samples). The near-zero global correlation therefore results from aggregating two groups with different baseline efficiency levels, which masks the underlying area-dependent scaling behavior.

\begin{figure}[ht]
  \centering
  \includegraphics[width=\columnwidth]{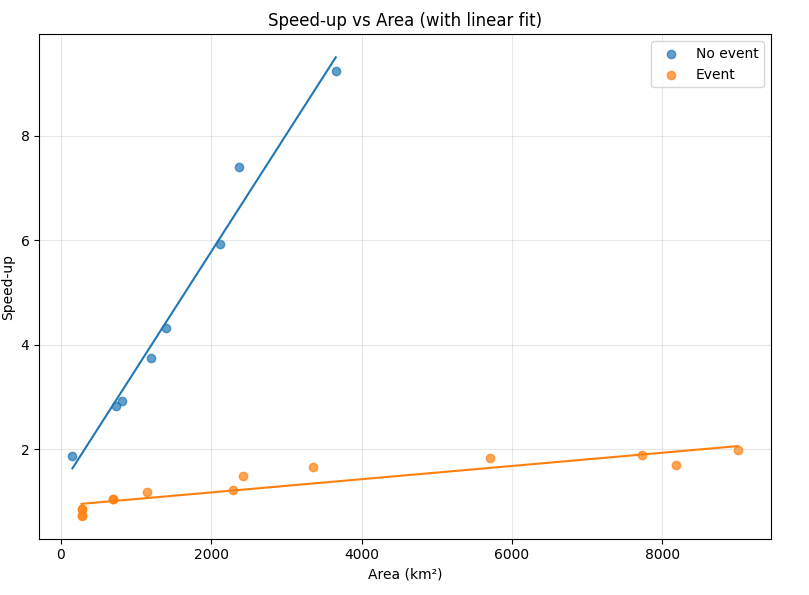}
  \caption{Scatter plot of speed-up versus scene area for event and non-event samples. Lines correspond to linear regression fits for each group.}
  \label{fig:speedup_vs_area}
\end{figure}

The relationship between speed-up and scene area within each regime is clearly illustrated in Fig.\ref{fig:speedup_vs_area}.

This behavior can be further explained by the computational characteristics of the involved components. In the baseline configuration, specialist tools rely in part on sliding-window processing strategies, which scale with the number of image patches and therefore increase computational cost proportionally to scene extent, with limited global parallelization. In contrast, the Early Warning stage leverages a vision-language model whose visual encoder processes the full image through convolutional operations, enabling more efficient parallelization across the spatial domain. As a result, the relative advantage of the routing strategy becomes more pronounced as scene size increases.

\subsection{Semantic explanation analysis}
While Section \ref{sec:performace} focused on computational efficiency, the proposed architecture also affects the qualitative characteristics of the generated explanations. In particular, the Early Warning Agent does not merely act as a routing mechanism, but introduces a structured prior hypothesis that guides downstream reasoning.

Beyond runtime improvements, the \emph{Early Warning} agent complements the specialists in two ways: it enables early stopping and selective routing for lower latency, while the specialists provide evidence to validate the initial hypothesis and reduce false positives. 

\begin{figure}[!t]
    \centering
    \begin{tcolorbox}[title=Example: Specialist refutes the Early Warning prediction]
    \fontsize{10}{12}\selectfont
    \textbf{Analyzed scene}

    \begin{center}
        \begin{minipage}{0.48\linewidth}
            \centering
            \includegraphics[width=\linewidth]{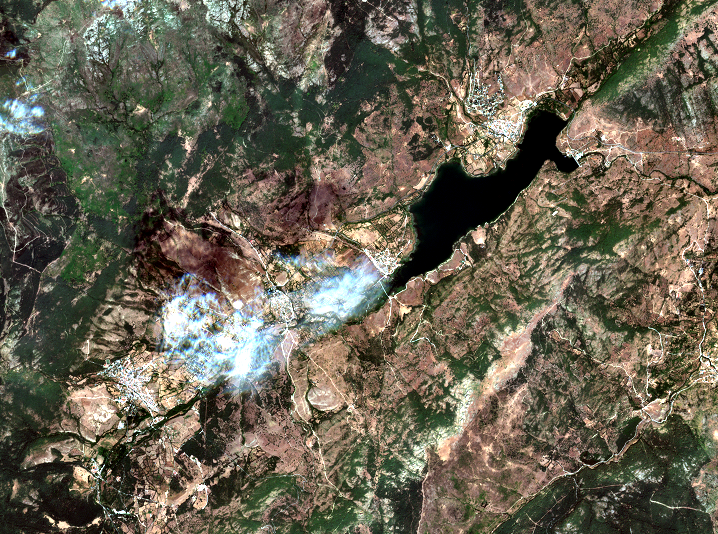}
            \small (a) RGB
        \end{minipage}
        \hfill
        \begin{minipage}{0.48\linewidth}
            \centering
            \includegraphics[width=\linewidth]{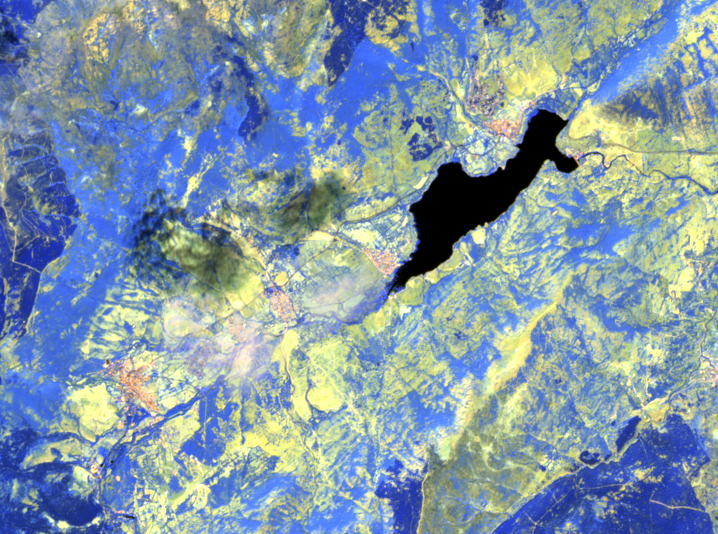}
            \small (b) SWIR false-color
        \end{minipage}
    \end{center}

    \medskip
    \hrule
    \medskip

    \textsc{\textbf{Early Warning Agent}}
    \smallskip

    \textbf{Predicted event:} \hspace{0.15em}\underline{Wildfire} \\
    \textbf{Reasoning:} \hspace{0.15em}The image suggests a potential wildfire in a mountainous region.

    \medskip
    \centerline{$\downarrow$}
    \medskip

    \textsc{\textbf{Wildfire Specialist Agent}}
    \smallskip

    \textbf{Reasoning:} \hspace{0.15em}All tools report \underline{no fire}:
    \begin{itemize}[noitemsep]
    \item ML fire detector: $0.0~\text{km}^2$
    \item SWIR index-based active fire: $0.0~\text{km}^2$
    \item Burned area estimate: $0$ hotspots, $0.0~\text{km}^2$
    \end{itemize}

    \medskip
    \centerline{$\downarrow$}
    \medskip

    \textsc{\textbf{Decision Agent}}
    \smallskip

    \textbf{Final decision:}\hspace{0.15em} \underline{No alert}

    \textbf{Reasoning:} \hspace{0.15em}Specialist reports using multiple detection tools found no fire, with all reports indicating $0.0~\text{km}^2$ of fire area. The Early Warning prediction did not align with the tool outputs, suggesting an anomaly or error in the prediction model.

    \end{tcolorbox}

    \caption{Example where specialist evidence contradicts the Early Warning prediction. The scene is shown using Sentinel-2 RGB and SWIR false-color composites (B12, B11, B8). The wildfire specialist finds no supporting evidence, leading the decision agent to reject the alert.}

    \label{fig:refute_example}

\end{figure}

This validation mechanism becomes particularly relevant when the initial hypothesis is incorrect. For instance, as illustrated in Fig.~\ref{fig:refute_example}, the Early Warning Agent predicts a wildfire, but the routed specialist tools find no supporting evidence
and the Decision Agent rejects the alert. This example highlights the internal consistency checks enabled by the layered architecture: rather than committing to the initial prediction, the system incorporates specialist evidence to confirm or invalidate the preliminary hypothesis.

Beyond this consistency check, we further analyze the qualitative properties of the explanations generated by both configurations. The baseline aggregates the outputs of both specialist agents into a single report, whereas the proposed pipeline produces explanations conditioned on the Early Warning hypothesis and the selectively activated
specialist.

\begin{figure}[!t]
    \centering
    \begin{tcolorbox}[title=Example: Wildfire Report]
    \fontsize{10}{12}\selectfont
    \textbf{Analyzed scene}

    \begin{center}
        \begin{minipage}{0.48\linewidth}
            \centering
            \includegraphics[width=\linewidth]{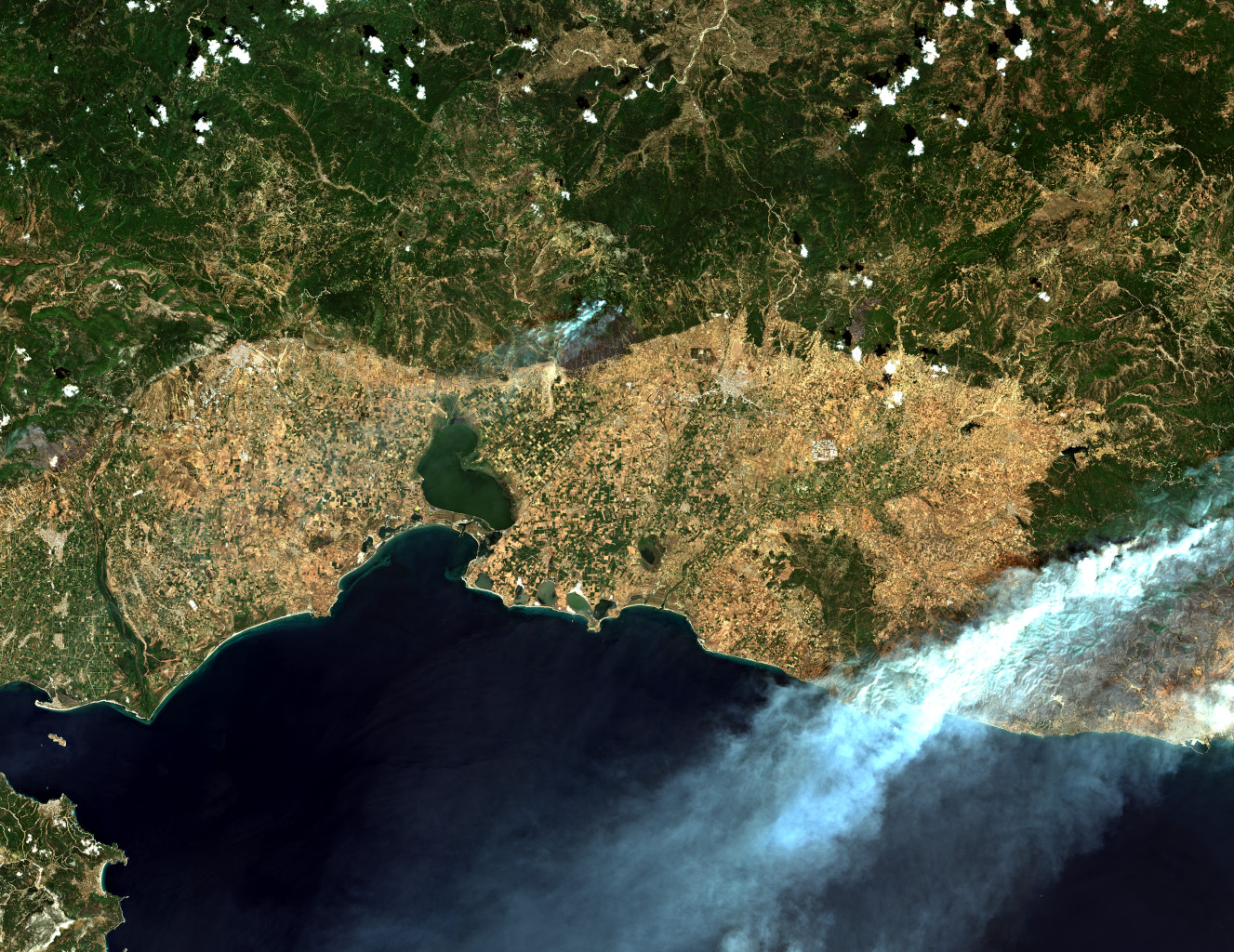}
            \small (a) RGB
        \end{minipage}
        \hfill
        \begin{minipage}{0.48\linewidth}
            \centering
            \includegraphics[width=\linewidth]{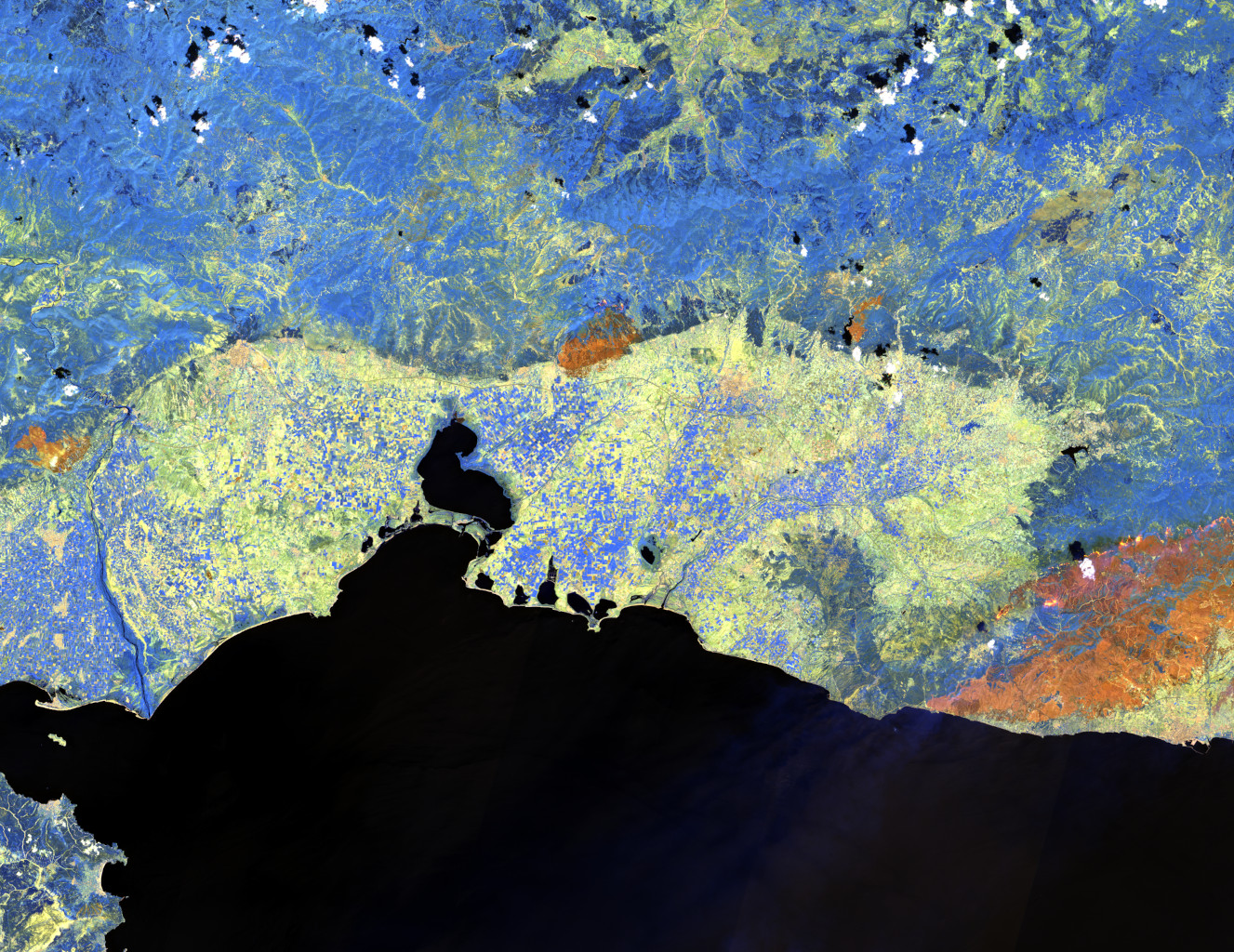}
            \small (b) SWIR false-color
        \end{minipage}
    \end{center}

    \medskip
    \hrule
    \medskip

\begin{center}
\begin{minipage}[t]{0.48\linewidth}

\textbf{Baseline}
\medskip
\medskip

\textsc{\textbf{Decision Agent}}

\smallskip
\textbf{Reasoning:}\hspace{0.15em}
The wildfire specialist confirms a fire detected with multiple tools
(ML and spectral indexes) and provides metrics on burned area.

\end{minipage}
\hfill
\begin{minipage}[t]{0.48\linewidth}

\textbf{Proposed pipeline}
\medskip
\medskip

\textsc{\textbf{Early Warning\\[-2pt]Agent}}

\smallskip

\textbf{Predicted event:}\underline{Wildfire}

\textbf{Reasoning:}\hspace{0.15em}
The image shows a wildfire spreading across a coastal area,
indicating a potential risk to the nearby population and infrastructure.

\medskip
\medskip

\textsc{\textbf{Decision Agent}}

\smallskip

\textbf{Reasoning:}\hspace{0.15em}
Specialist reports confirm the detection of a wildfire with active fire
area of $1.77~\text{km}^2$, and an estimated burned area of
$41.1~\text{km}^2$. The reports align with the early-warning report
indicating a potential risk to the nearby population and infrastructure.

\end{minipage}

\end{center}

\end{tcolorbox}

\caption{Comparison of wildfire reports generated by the baseline configuration (two specialists and a decision agent) and the proposed hierarchical pipeline. The scene is shown using Sentinel-2 RGB and SWIR false-color composites (B12, B11, B8). Both configurations detect the wildfire, but the proposed pipeline provides a more focused explanation aligned with the routed specialist evidence.}

\label{fig:wildfire_report1}

\end{figure}

\begin{figure}[!t]
    \centering
    \begin{tcolorbox}[title=Example: Flood Report]
    \fontsize{10}{12}\selectfont
    \textbf{Analyzed scene}

    \begin{center}
        \begin{minipage}{0.48\linewidth}
            \centering
            \includegraphics[width=\linewidth]{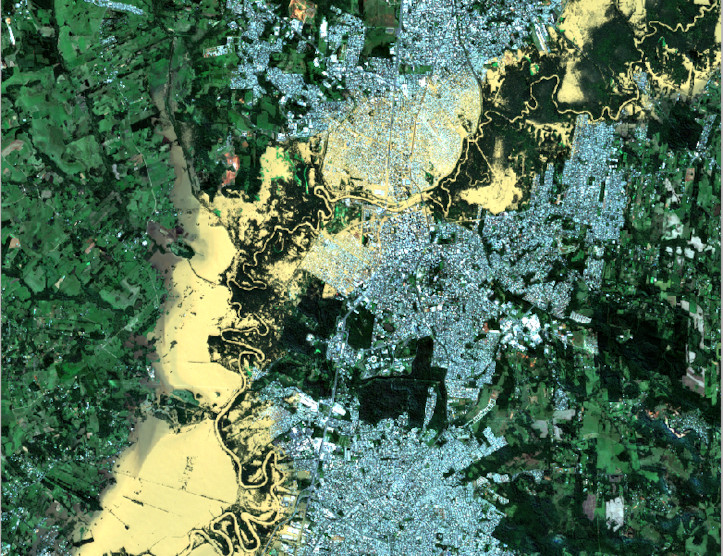}
            \small (a) RGB
        \end{minipage}
        \hfill
        \begin{minipage}{0.48\linewidth}
            \centering
            \includegraphics[width=\linewidth]{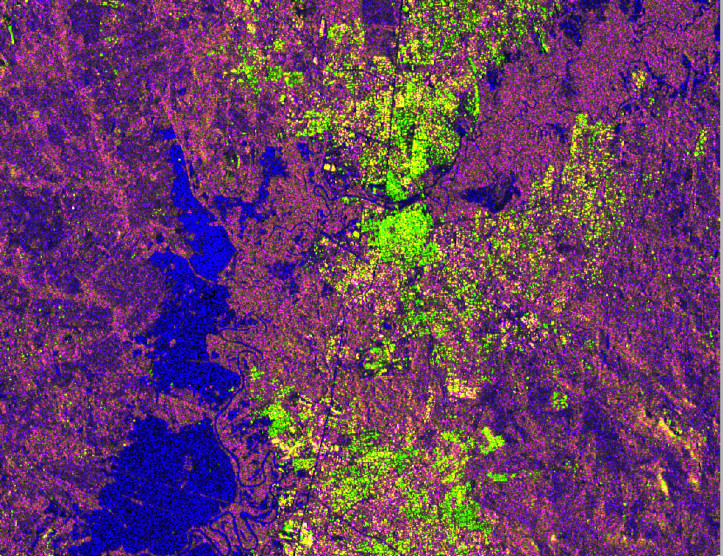}
            \small (b) SAR false-color
        \end{minipage}
    \end{center}

    \medskip
    \hrule
    \medskip

    \begin{center}
        \begin{minipage}[t]{0.48\linewidth}

            \textbf{Baseline}
            \medskip
            \medskip

            \textsc{\textbf{Decision Agent}}

            \smallskip
            \textbf{Reasoning:}\hspace{0.15em}
            The flood specialist clearly confirms a flood with an area of $25.42~\text{km}^2$.

        \end{minipage}
        \hfill
        \begin{minipage}[t]{0.48\linewidth}

            \textbf{Proposed pipeline}
            \medskip
            \medskip

            \textsc{\textbf{Early Warning\\[-2pt]Agent}}

            \smallskip

            \textbf{Predicted event:}\hspace{0.3em}\underline{Flood}

            \textbf{Reasoning:}\hspace{0.15em}
            The image shows a flooded area with orange colors, indicating a flood. The city is surrounded by floodwaters, which poses significant risks to infrastructure and human life.

            \medskip
            \medskip

            \textsc{\textbf{Decision Agent}}

            \smallskip

            \textbf{Reasoning:}\hspace{0.15em}
            Specialist reports confirm the presence of a flood with an area of $25.42~\text{km}^2$. This indicates significant risks to the city's infrastructure and poses threats to human life.

        \end{minipage}

    \end{center}

    \end{tcolorbox}

    \caption{Comparison of flood reports generated by the baseline configuration (two specialists and a decision agent) and the proposed hierarchical pipeline. The scene is shown using Sentinel-2 RGB and SAR false-color composites (VV, VH, VV/VH). Both detect the flood, but the proposed pipeline provides a richer semantic explanation of the event and its potential impact.}
    \label{fig:flood_report1}

\end{figure}

\begin{figure}[!t]
    \centering
    \begin{tcolorbox}[title=Example: Past Burn vs Active Wildfire]
    \fontsize{10}{12}\selectfont
    \textbf{Analyzed scene}

    \begin{center}
        \begin{minipage}{0.48\linewidth}
            \centering
            \includegraphics[width=\linewidth]{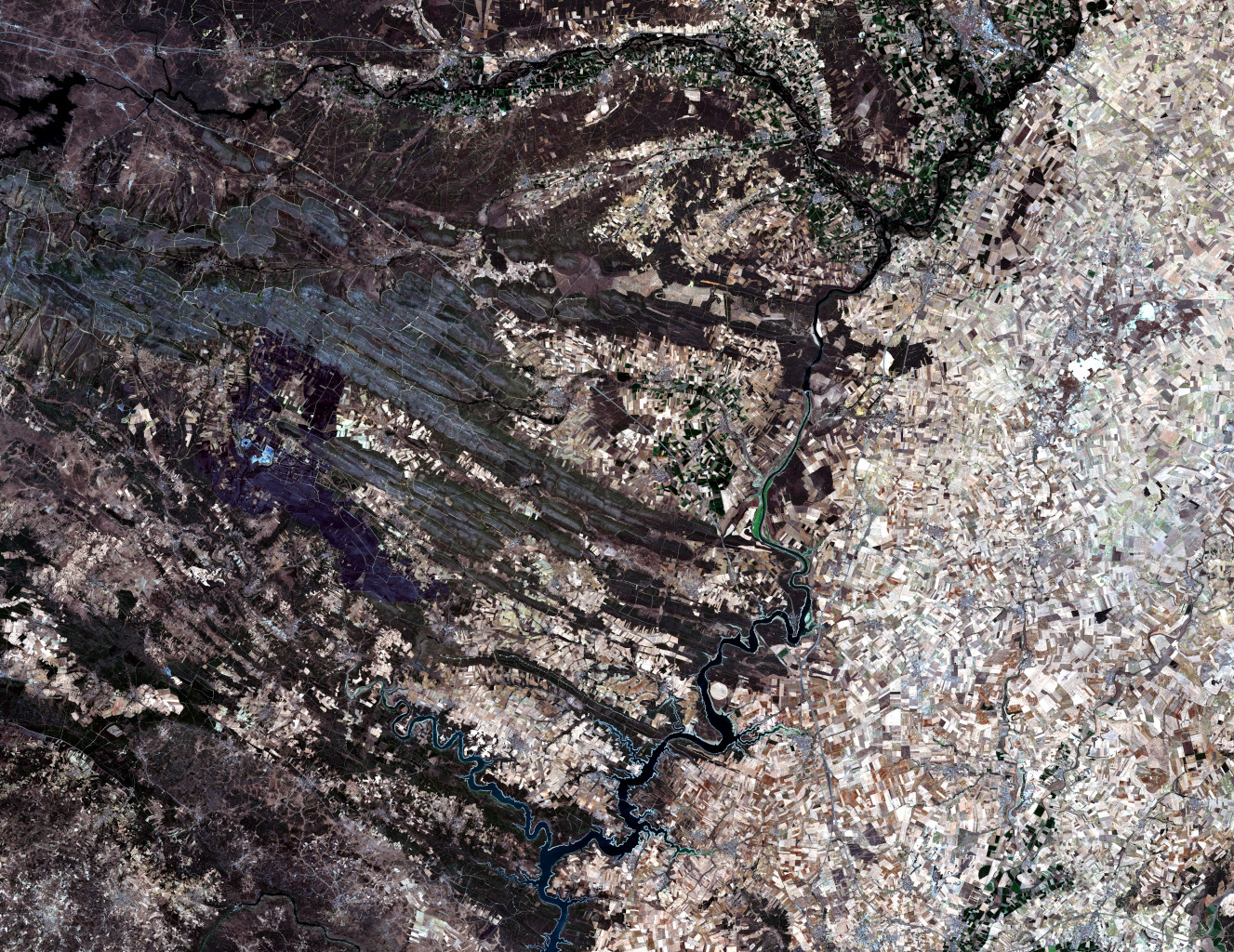}
            \small (a) RGB
        \end{minipage}
        \hfill
        \begin{minipage}{0.48\linewidth}
            \centering
            \includegraphics[width=\linewidth]{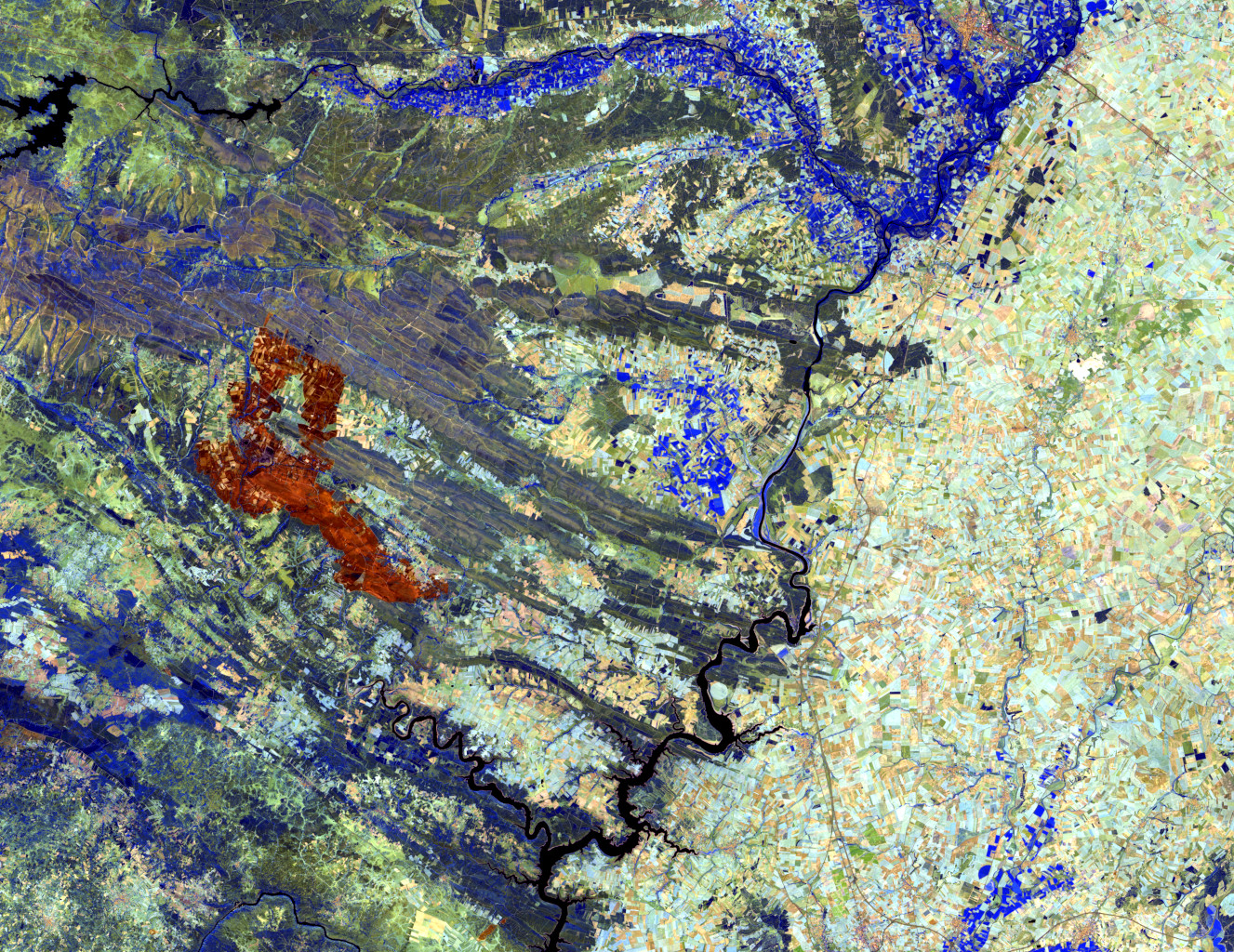}
            \small (b) SWIR false-color
        \end{minipage}
    \end{center}

    \medskip
    \hrule
    \medskip

    \begin{center}
        \begin{minipage}[t]{0.48\linewidth}

            \textbf{Baseline}
            \medskip
            \medskip

            \textsc{\textbf{Decision Agent}}

            \smallskip
            \textbf{Reasoning:}\hspace{0.15em}
            The wildfire specialist detected a burned area of $44.83~\text{km}^2$ suggesting a possible wildfire.

        \end{minipage}
        \hfill
        \begin{minipage}[t]{0.48\linewidth}

            \textbf{Proposed pipeline}
            \medskip
            \medskip

            \textsc{\textbf{Early Warning\\[-2pt]Agent}}

            \smallskip

            \textbf{Predicted event:}\underline{Wildfire}

            \textbf{Reasoning:}\hspace{0.15em}
            It seems there is a bit of smoke, typical of a wildfire

            \medskip
            \medskip

            \textsc{\textbf{Decision Agent}}

            \smallskip

            \textbf{Reasoning:}\hspace{0.15em}
            Specialist reports an estimated burned area of $44.83~\text{km}^2$, suggesting past wildfires, but no new active fires were identified.

        \end{minipage}

    \end{center}

    \end{tcolorbox}

    \caption{Comparison between the baseline configuration (two specialists and a decision agent) and the proposed hierarchical pipeline in a scene containing previously burned areas. The Sentinel-2 scene is shown using RGB and SWIR false-color composites (B12, B11, B8). The baseline interprets the burn scar as an active wildfire, whereas the proposed pipeline correctly identifies it as a past event with no active fire.}
    
    \label{fig:past_fire}

\end{figure}

The following examples highlight three complementary aspects of the proposed architecture: hypothesis validation, explanation grounding, and improved semantic interpretation. These qualitative differences between both approaches are illustrated in Figs.~\ref{fig:wildfire_report1}–\ref{fig:past_fire}.

In Fig.~\ref{fig:wildfire_report1}, both configurations correctly identify the wildfire event. However, the proposed pipeline produces a more focused explanation that is explicitly grounded in the routed specialist evidence.

Similarly, the example in Fig.~\ref{fig:flood_report1} shows how the proposed pipeline combines the specialist’s quantitative estimate with a clearer semantic interpretation of the consequences in a populated area, highlighting potential risks beyond simple detection.

A different qualitative behavior is illustrated in Fig.~\ref{fig:past_fire}. In this case, the baseline interprets the burned area as a potential ongoing wildfire, whereas the proposed
pipeline correctly identifies it as a past burn event with no active fire.

Across these examples, the proposed pipeline consistently produces explanations that are both more focused and more context-aware. The layered reasoning structure therefore improves semantic coherence without increasing computational cost.

\section{General Discussion}
\label{sec:discussion}

The results highlight the effectiveness of the proposed architecture, particularly in reducing latency through early-stage routing while maintaining coherent decision outputs. This advantage is especially pronounced in non-disaster scenarios, which are expected to dominate in realistic operational settings, suggesting that the practical efficiency gains could be even greater in real-world deployments.

These reductions in processing time also have implications for energy efficiency. Since power consumption remains relatively stable, reduced processing time translates into lower energy usage. More importantly, it increases the availability of specialist resources, allowing more candidate events to be analyzed and improving detection reliability.

In scenes containing actual disasters, the speed-up is less pronounced, particularly when the affected area is relatively small. This behavior can be largely explained by the additional latency introduced by the Early Warning VLM. In principle, this component could be replaced by a much lighter classifier, which would preserve the routing capability while making the difference with respect to the baseline even more evident. However, the purpose of using a VLM at the first node is not limited to demonstrating the benefits of routing alone. Its role is also to provide high-level semantic information that can assist the decision agent in producing a more informed and coherent final verdict, which constitutes a key advantage of the proposed architecture beyond raw performance gains.

It is also important to note that the VLM used in the current demonstrator is a generic model, with no fine-tuning for satellite imagery and, even less, for the description of natural-disaster scenes. In addition, it only processes RGB imagery, which limits its applicability in Earth Observation context where other sensing modalities are often crucial. Recent reviews also highlight the limitations of generic models not specifically adapted to EO data, particularly when dealing with spectral products, SAR, and multi-sensor data \cite{agentic_ai_survey}. These limitations are expected at this stage and do not stem from the proposed architecture itself, and adapting this component to the EO domain would likely lead to a significant improvement in both routing and the semantic quality of the final output.

The evaluation is based on a limited dataset and is intended to demonstrate the functional behavior of the proposed architecture rather than provide a comprehensive performance benchmark. Despite this, the results consistently illustrate the expected system behavior. The current implementation should therefore be interpreted as a proof-of-concept.

An additional factor to consider is the hardware configuration used during the experiments. The experimental evaluation was conducted on a space-qualified CPU platform without dedicated AI acceleration. Dedicated inference hardware would significantly reduce the latency of the main computational components of the system, including the early-warning vision–language model, the deep-learning vision models used by the specialist agents, and the language models responsible for evidence interpretation and decision fusion. Beyond improving processing speed, such accelerators would also enable the deployment of more capable language models, potentially allowing deeper reasoning over multimodal evidence and improving the semantic quality of the final decision outputs. Even under these constraints, the system already demonstrates the semantic value of the proposed approach, suggesting that the reported performance represents a conservative estimate of its potential.

On the other hand, the specialist agents use their tools in a predefined manner. They do not decide autonomously which analysis to apply or when to apply it; instead, the sequence of operations is fixed and ordered. Allowing them full freedom to choose the required analyses led to repeated tool calls and, in some cases, even looping behavior. This behavior is consistent with recent findings describing tool orchestration in remote-sensing agents as fragile, with frequent issues in tool selection and repeated unsuccessful calls \cite{agentic_ai_survey}. These observations are also reflected in recent benchmarks showing that LLM agents frequently fail when required to autonomously generate EO analysis workflows over platforms such as Google Earth Engine \cite{univearth}, supporting the design choice of structured and constrained agent roles in the proposed architecture.

The purpose of this work does not lie in the specific topology adopted. The Early Warning agent could, in principle, handle a broader range of possibilities: rather than being limited to either discarding an observation or unilaterally routing it to a specialist, it could request a second opinion even under mild suspicion. Likewise, a node could trigger an additional inspection of a specific area, request a more detailed analysis, or even seek an alternative perspective from another satellite, possibly from a different viewing angle or acquisition time in order to assess the evolution of the scene, highlighting the extensibility of the proposed framework.

Nor does the contribution depend on the particular tools employed. As discussed previously, the VLM used in the Early Warning stage could be replaced by a lightweight classifier; the downstream analyses could rely on different methods; and the agents themselves could be built on other LLMs, or even on specialized vision language models. The objective of this work is to explore and demonstrate the advantages of a hierarchical and distributed architecture based on early detection, functional specialization, and evidence-driven interpretable decision-making, independently of the specific implementation choices.

Finally, if multiple early-warning nodes were deployed, most candidate detections would still correspond to benign scenarios. However, each of these candidates could trigger requests to specialized analysis modules, potentially saturating intermediate processing resources. In such settings, mechanisms for prioritizing observations and arbitrating resource allocation become necessary. Approaches based on distributed decision-making for satellite constellations, such as the multi-agent federation strategies proposed by \cite{federation}, could provide a framework for coordinating these decisions across multiple assets, representing a promising direction for extending the proposed architecture toward fully distributed operational scenarios.

\section{Conclusions and Future Work}
\label{sec:conclusions}
This work presented a hierarchical multi-agent architecture for onboard Earth Observation processing of multimodal observations, enabling early hypothesis generation, role-specialized analysis, and decision-level evidence fusion. A proof-of-concept implementation executed on the engineering model of a representative onboard edge-computing platform demonstrates that early-stage hypothesis generation can significantly reduce unnecessary computation while maintaining coherent and interpretable decision outputs, with improved semantic consistency.

These results highlight the potential of structured, agent-based reasoning to move beyond traditional ground-centric processing pipelines, enabling more efficient and responsive onboard decision-making in future EO systems. By selectively allocating computational resources based on event likelihood, the proposed approach provides a practical pathway toward scalable and energy-efficient onboard intelligence under realistic resource constraints.

More broadly, this work supports the feasibility of distributed, role-specialized processing architectures as a foundation for next-generation autonomous EO missions. Future research may explore more advanced coordination mechanisms between agents, improved adaptation of language and vision–language models to EO data, and the integration of richer multimodal observations to further enhance system robustness and interpretability.


\bibliographystyle{unsrt}
\bibliography{chapters/bibliography}

@article{astrea,
  title={ASTREA: Introducing agentic intelligence for orbital thermal autonomy},
  author={Mousist, Alejandro D},
  journal={arXiv preprint arXiv:2509.13380},
  year={2025}
}

@article{agentic_ai_survey,
  title={Agentic AI in Remote Sensing: Foundations, Taxonomy, and Emerging Systems},
  author={Talemi, Niloufar Alipour and Boone, Julia and Afghah, Fatemeh},
  journal={arXiv preprint arXiv:2601.01891},
  year={2026}
}

@article{earth-agent,
  title={Earth-agent: Unlocking the full landscape of earth observation with agents},
  author={Feng, Peilin and Lv, Zhutao and Ye, Junyan and Wang, Xiaolei and Huo, Xinjie and Yu, Jinhua and Xu, Wanghan and Zhang, Wenlong and Bai, Lei and He, Conghui and others},
  journal={arXiv preprint arXiv:2509.23141},
  year={2025}
}

@inproceedings{naiad,
  title={Naiad: novel agentic intelligent autonomous system for inland water monitoring},
  author={Baltzi, Eirini and Moumouris, Tilemachos and Psalta, Athina and Tsironis, Vasileios and Karantzalos, Konstantinos},
  booktitle={Remote Sensing for Agriculture, Ecosystems, and Hydrology XXVII},
  volume={13666},
  pages={187--198},
  year={2025},
  organization={SPIE}
}

@misc{rs-agent,
      title={RS-Agent: Automating Remote Sensing Tasks through Intelligent Agent}, 
      author={Wenjia Xu and Zijian Yu and Boyang Mu and Zhiwei Wei and Yuanben Zhang and Guangzuo Li and Jiuniu Wang and Mugen Peng},
      year={2026},
      eprint={2406.07089},
      archivePrefix={arXiv},
      primaryClass={cs.CV},
      url={https://arxiv.org/abs/2406.07089}, 
}

@article{multiagent,
  title={Autonomous Agents and Multiagent Systems Challenges in Earth Observation Satellite Constellations},
  author={Picard, Gauthier and Caron, Cl{\'e}ment and Farges, Jean-Loup and Guerra, Jonathan and Pralet, C{\'e}dric and Roussel, St{\'e}phanie},
  year={2021}
}

@inproceedings{federation,
  title={Going beyond mono-mission earth observation: Using the multi-agent paradigm to federate multiple missions},
  author={Farges, Jean-Loup and Perotto, Filipo Studzinski and Pralet, C{\'e}dric and Picard, Gauthier and de Lussy, Cyril and Guerra, Jonathan and Pavero, Philippe and Planchou, Fabrice},
  booktitle={23rd International Conference on Autonomous Agents and Multiagent Systems (AAMAS-24)},
  pages={2674--2678},
  year={2024},
  organization={International Foundation for Autonomous Agents and Multiagent Systems}
}

@article{eo-alert,
  title={Novel Operational Scenarios for the Next-Generation Earth Observation Satellites Supporting On-Board Processing for Rapid Civil Alerts S. Cornara1*, S. Tonetti1, M. Kerr1, G. Vicario de Miguel1, S. Fraile2, M. D{\'\i}ez2, H. Breit3, S. Wiehle3, C.},
  author={Mart{\'\i}n, Marcos and Solimini, C},
  year={2021}
}

@misc{univearth,
      title={Towards LLM Agents for Earth Observation}, 
      author={Chia Hsiang Kao and Wenting Zhao and Shreelekha Revankar and Samuel Speas and Snehal Bhagat and Rajeev Datta and Cheng Perng Phoo and Utkarsh Mall and Carl Vondrick and Kavita Bala and Bharath Hariharan},
      year={2025},
      eprint={2504.12110},
      archivePrefix={arXiv},
      primaryClass={cs.AI},
      url={https://arxiv.org/abs/2504.12110}, 
}

@misc{geollm_squad,
      title={Multi-Agent Geospatial Copilots for Remote Sensing Workflows}, 
      author={Chaehong Lee and Varatheepan Paramanayakam and Andreas Karatzas and Yanan Jian and Michael Fore and Heming Liao and Fuxun Yu and Ruopu Li and Iraklis Anagnostopoulos and Dimitrios Stamoulis},
      year={2025},
      eprint={2501.16254},
      archivePrefix={arXiv},
      primaryClass={cs.LG},
      url={https://arxiv.org/abs/2501.16254}, 
}

@inproceedings{starling,
  title={Smallsat 2024-Starling Cubesat Swarm Technology Demonstration Flight Results},
  author={Miller, Scott and Adams, Caleb and Alem, Nahum and Cannon, Howard and Grashuis, Randy and Hendriks, Ted and Hwang, Soon and Iatauro, Michael and Pires, Craig and Kruger, Justin and others},
  booktitle={Proceedings of the 38th Annual Small Satellite Conference},
  number={SSC24-I-06},
  year={2024},
  organization={Small Satellite Conference}
}

@article{iride,
  title={IRIDE, the Euro-Italian earth observation program: overview, current progress, global expectations, and recommendations},
  author={Orusa, Tommaso and Viani, Annalisa and Borgogno-Mondino, Enrico},
  journal={Environmental Sciences Proceedings},
  volume={29},
  number={1},
  pages={74},
  year={2024},
  publisher={MDPI}
}

@article{atlantic_constellation,
  title={The Atlantic Constellation Very High Resolution, A Small Satellite Approach Achieving High-Performance Optical Imagery},
  author={Candeias, Henrique and Rold{\'a}n, Carlos and Velarde, Carmen and Almeida, Filipe and Ferreira, Francisco Miguel and Aguiar, Gon{\c{c}}alo and Correia, Gon{\c{c}}alo and Santos, Henrique and Leit{\~a}o, Jo{\~a}o Pedro and Meyer, John and others},
  year={2025}
}

@misc{qwen2.5,
    title = {Qwen2.5: A Party of Foundation Models},
    url = {https://qwenlm.github.io/blog/qwen2.5/},
    author = {Qwen Team},
    month = {September},
    year = {2024}
}

@article{qwen2vl,
  title={Qwen2-vl: Enhancing vision-language model's perception of the world at any resolution},
  author={Wang, Peng and Bai, Shuai and Tan, Sinan and Wang, Shijie and Fan, Zhihao and Bai, Jinze and Chen, Keqin and Liu, Xuejing and Wang, Jialin and Ge, Wenbin and others},
  journal={arXiv preprint arXiv:2409.12191},
  year={2024}
}

@inproceedings{IMAGINe,
  author    = {Elisa Callejo and Lara Arche and Alejandro Mousist and Adam Loverro},
  title     = {IMAGIN-e: The First Step Towards Extending the Cloud into Space},
  booktitle = {Proceedings of the 2023 Conference on Big Data from Space (BiDS’23)},
  year      = {2023},
  pages     = {397--400},
  publisher = {Publications Office of the European Union},
  address   = {Vienna, Austria},
  month     = {November},
  doi       = {10.2760/46796},
  url       = {https://op.europa.eu/en/publication-detail/-/publication/10ba86b1-7c63-11ee-99ba-01aa75ed71a1/language-en}
}

@misc{prefect_open_source,
  author       = {Prefect Technologies, Inc.},
  title        = {Prefect - Open Source Workflow Orchestration},
  howpublished = {\url{https://www.prefect.io/prefect/open-source}},
  note         = {Accessed: 2026-02-11},
}

@ARTICLE{nhi,
  author={Mazzeo, Giuseppe and Falconieri, Alfredo and Filizzola, Carolina and Genzano, Nicola and Pergola, Nicola and Marchese, Francesco},
  journal={IEEE Transactions on Geoscience and Remote Sensing}, 
  title={Wildfire Detection and Mapping by Satellite With an Enhanced Configuration of the Normalized Hotspot Indices: Results From Sentinel-2 and Landsat 8/9 Data Integration}, 
  year={2025},
  volume={63},
  number={},
  pages={1-21},
  doi={10.1109/TGRS.2025.3528641}}

@article{mndwi,
  title={Modification of normalised difference water index (NDWI) to enhance open water features in remotely sensed imagery},
  author={Xu, Hanqiu},
  journal={International journal of remote sensing},
  volume={27},
  number={14},
  pages={3025--3033},
  year={2006},
  publisher={Taylor \& Francis}
}

@inproceedings{deeplabv3plus,
  title={Encoder-decoder with atrous separable convolution for semantic image segmentation},
  author={Chen, Liang-Chieh and Zhu, Yukun and Papandreou, George and Schroff, Florian and Adam, Hartwig},
  booktitle={Proceedings of the European conference on computer vision (ECCV)},
  pages={801--818},
  year={2018}
}

@InProceedings{resnet50,
author = {He, Kaiming and Zhang, Xiangyu and Ren, Shaoqing and Sun, Jian},
title = {Deep Residual Learning for Image Recognition},
booktitle = {Proceedings of the IEEE Conference on Computer Vision and Pattern Recognition (CVPR)},
month = {June},
year = {2016}
}

@article{senforflood,
  title={SenForFlood: A New Global Dataset for Flooded Area Detection},
  author={Matosak, Bruno Menini and Gella, Getachew Workineh and Lang, Stefan},
  journal={The International Archives of the Photogrammetry, Remote Sensing and Spatial Information Sciences},
  volume={48},
  pages={97--102},
  year={2025},
  publisher={Copernicus Publications G{\"o}ttingen, Germany}
}

@article{transfer_learning_tf,
  title={Active fire segmentation: A transfer learning study from Landsat-8 to Sentinel-2},
  author={Fusioka, Andr{\'e} Minoro and de Almeida Pereira, Gabriel Henrique and Nassu, Bogdan Tomoyuki and Minetto, Rodrigo},
  journal={IEEE Journal of Selected Topics in Applied Earth Observations and Remote Sensing},
  year={2024},
  publisher={IEEE}
}

@inproceedings{sen2fire,
  title={Sen2fire: A challenging benchmark dataset for wildfire detection using sentinel data},
  author={Xu, Yonghao and Berg, Amanda and Haglund, Leif},
  booktitle={IGARSS 2024-2024 IEEE International Geoscience and Remote Sensing Symposium},
  pages={239--243},
  year={2024},
  organization={IEEE}
}

\small

\end{document}